\documentclass[10pt,twocolumn,letterpaper]{article}

\usepackage[pagenumbers]{cvpr} 

\usepackage{graphicx}
\usepackage{amsmath}
\usepackage{amssymb}
\usepackage{booktabs}

\usepackage{times}
\usepackage{epsfig}

\usepackage{float}

\usepackage[table]{xcolor}

\usepackage{footmisc}
\usepackage[pagebackref=false,breaklinks=true,colorlinks,bookmarks=false]{hyperref}
\usepackage{cleveref}
 \usepackage[verbose]{placeins} 
 \usepackage{multirow}
 \usepackage{enumitem}

\newcommand\Mark[1]{\textsuperscript#1}

\begin{document}

\title{
\vspace{-8mm}DifFIQA: Face Image Quality Assessment Using Denoising Diffusion Probabilistic Models
\vspace{-4mm}}

\author{Žiga Babnik\Mark{1}, Peter Peer\Mark{2}, Vitomir Štruc\Mark{1}\\
\Mark{1}University of Ljubljana, Faculty of Electrical Engineering\\ 
\Mark{2}University of Ljubljana, Faculty of Computer and Information Science\\
{\tt\small \{ziga.babnik, vitomir.struc\}@fe.uni-lj.si, peter.peer@fri.uni-lj.si}
}
\twocolumn[{%
\renewcommand\twocolumn[1][]{#1}%
\maketitle
\begin{center}
    \centering
    \vspace{-8mm}
    \captionsetup{type=figure}
    \includegraphics[width=.9\textwidth, trim = 7mm 0 0 0, clip]{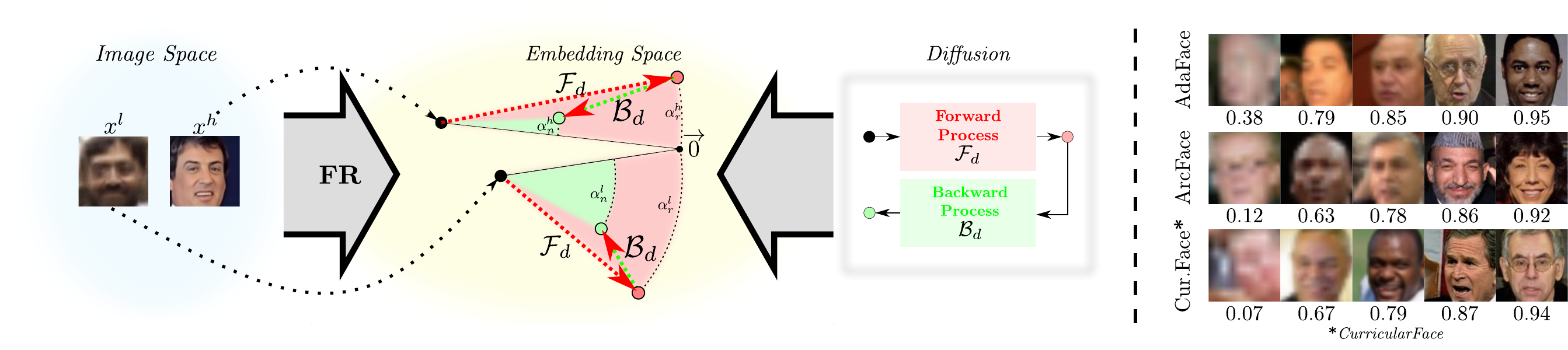}\vspace{-4mm}
    \captionof{figure}{\textbf{High-level idea behind the proposed DifFIQA face image quality assessment (FIQA) approach.} The quality of face images corresponds to a considerable degree to the stability of the respective representations in the embedding space of a given face recognition~(FR) model. DifFIQA utilizes a diffusion framework to explore the embedding stability through image perturbations caused by the noising and denoising processes. The intuition behind this approach is that the forward~(noising) $\mathcal{F}_d$ and backward~(denoising) $\mathcal{B}_d$  diffusion processes lead to larger embedding perturbations for lower-quality images ($x^l$) compared to facial images of higher quality ($x^h$).  
    By analyzing the impact of both the forward and backward processes on the representation of a given image, DifFIQA is able to infer the corresponding quality and/or generate (FR model specific) quality rankings, as shown on the right. The figure is best viewed electronically.\vspace{3mm}}
    \label{fig:teaser}
\end{center}%
}]

\begin{abstract}\vspace{-3mm}
Modern face recognition (FR) models excel in constrained scenarios, but often suffer from decreased performance when deployed in unconstrained (real-world) environments due to uncertainties surrounding the quality of the captured facial data. Face image quality assessment (FIQA) techniques aim to mitigate these performance degradations by providing FR models with sample-quality predictions that can be used to reject low-quality samples and reduce false match errors. However, despite steady improvements, ensuring reliable quality estimates across facial images with diverse characteristics remains challenging. In this paper, we present a powerful new FIQA approach, named DifFIQA, which relies on  denoising diffusion probabilistic models (DDPM) and ensures highly competitive results. The main idea behind the approach is to utilize the forward and backward processes of DDPMs to perturb facial images and quantify the impact of these perturbations on the corresponding image embeddings for quality prediction. Because the diffusion-based perturbations are computationally expensive, we also distill the knowledge encoded in DifFIQA into a regression-based quality predictor, called DifFIQA(R), that balances performance and execution time. We evaluate both models in comprehensive experiments on 7 datasets, with 4 target FR models and against 10 state-of-the-art FIQA techniques with highly encouraging results. The source code will be made publicly available.
\end{abstract}

\section{Introduction}\label{sec:introduction}


State-of-the-art face recognition (FR) models achieve near-perfect results on various benchmarks with high-quality facial images, but still struggle in real-world situations, where the quality of the input samples is frequently unknown~\cite{fr1, fr2,grm2018strengths}. For instance, surveillance, a common application of FR, often involves lower quality samples due to unconstrained and covert capture conditions. In such cases, assessing the quality of the face-image samples is crucial. Low-quality samples can mislead the FR models and cause catastrophic false-match errors, leading to privacy breaches or even monetary loss. By determining the quality of input samples and rejecting or requesting recapture of those below a given quality threshold, the stability and performance of FR models can typically be improved.


Face Image Quality Assessment (FIQA) methods provide FR methods with a quality estimate for each given face sample. In this context, the term \textit{quality} can refer to either the character, fidelity or utility of the sample, as defined by ISO/IEC 29794-1~\cite{eval2}. Similarly to most FIQA research, we focus on the biometric utility of the facial samples, rather than the visual quality (character and fidelity) as perceived by humans \cite{survey}. Such image characteristics are commonly evaluated by general-purpose Image Quality Assessment (IQA) techniques. Biometric utility encompasses several unknown aspects of the given face sample, including its visual quality, face-specific information, and the relative biases inherent to the targeted FR model. It can be interpreted as the usefulness (or fitness) of the sample for the recognition task. Several types of FIQA techniques have been proposed over the years. The largest group focuses on training regression models from calculated pseudo reference quality labels~\cite{faceqnet, sdd-fiqa, pcnet, lightqnet}, with differences between methods in how they calculate the labels. Other approaches include (unsupervised) analytical methods~\cite{ser-fiq, faceqan, analytical2} that use reference-free approaches for quality prediction, 
and model-based solutions~\cite{pfe, magface, cr-fiqa} that combine the face recognition and quality assessment tasks. While modern FIQA techniques have demonstrated impressive performance, providing reliable quality predictions for diverse facial characteristics is still a challenging task.

In this paper, we introduce a novel FIQA technique, called DifFIQA (\textbf{Dif}fusion-based \textbf{F}ace \textbf{I}mage \textbf{Q}uality \textbf{A}ss\-essment), that leverages the image-generation versatility of modern Denoising Diffusion Probabilistic Models (DDPMs) for face quality assessment and generalizes well across a wide variety of datasets and FR models. As shown in Figure~\ref{fig:teaser}, DifFIQA is based on the following two insights:\vspace{-2mm}
\begin{itemize}[noitemsep]
    \item \textbf{Perturbation robustness:} Images of higher-quality have stable representations in the embedding space of the given FR model and are less effected by noise perturbations introduced by the forward diffusion process.
    \item \textbf{Reconstruction quality:} High-quality samples are easier to reconstruct from partially corrupted (noisy) data with incomplete identity information and exhibit less disparity between the embeddings of the input and denoised samples than low-quality images. 
\end{itemize}\vspace{-2mm}
Based on these observations, DifFIQA analyzes the embedding stability of the input image by perturbing it through the forward as well as backward diffusion process and infers a quality score from the result. To avoid the computationally expensive backward process and speed up computation, we also distill the DifFIQA approach into a regression-based model, termed DifFIQA(R). We evaluate both techniques through extensive experiments over multiple datasets and FR models, and show that both techniques lead to highly competitive results when compared to the state-of-the-art. 



\begin{figure*}[!ht]
    \centering
    \includegraphics[width=0.8\textwidth]{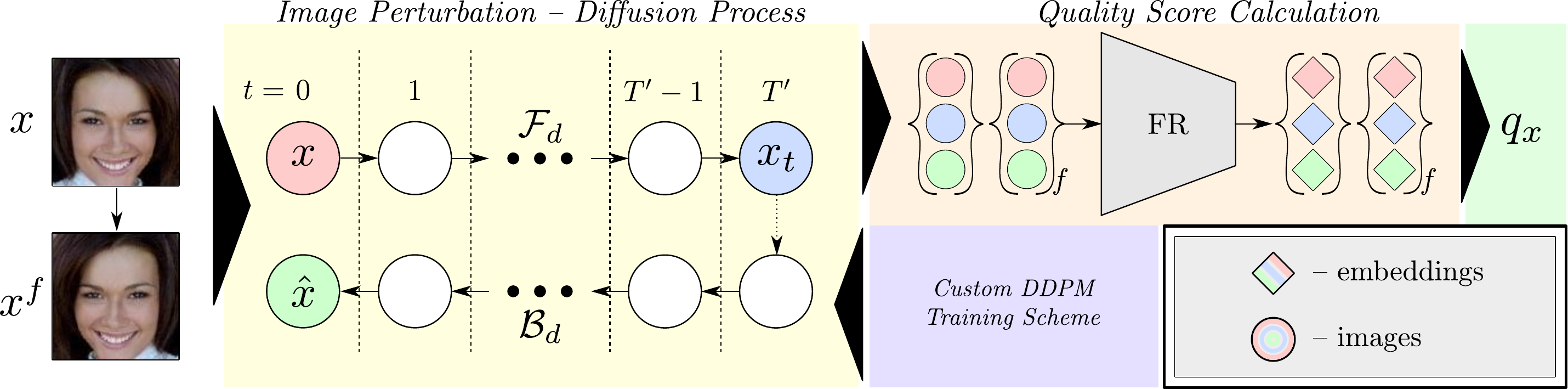}\vspace{0.5mm}
    \caption{\textbf{Overview of DifFIQA.} The proposed quality assessment pipeline consists of two main parts: the \textit{Diffusion Process} and the \textit{Quality-Score Calculation}. The diffusion process uses an encoder-decoder UNet model ($D$), trained using an extended DDPM training scheme that helps to generate higher-quality (restored) images. The custom DDPM model is used in the Diffusion Process, which generates noisy $x_t$ and reconstructed $\hat{x}$ images using the forward and backward diffusion processes, respectively. To capture the effect of facial pose on the quality estimation procedure, the process is repeated with a horizontally flipped image $x^f$. The Quality Score Calculation part then produces and compares the embeddings of the original images and the images generated by the diffusion part. 
    }
    \label{fig:overview}
\end{figure*}

\section{Related Work}\label{sec:related_work}

In this section, we briefly review existing FIQA solutions, which can conveniently be partitioned into three main groups: $(i)$ \textit{analytical} techniques, $(ii)$ \textit{regression-based} approaches, and $(iii)$ \textit{model-based} methods. \vspace{0.8mm} 

\noindent\textbf{Analytical methods.} The vast majority of methods from this group can be viewed as \textit{specialized} general-purpose IQA techniques that focus on quality predictions defined by $(i)$ selected visual characteristics of faces, such as pose, symmetry or interocular distance, and/or $(ii)$ general visual image properties, such as sharpness, illumination or noise. An early method from this group was presented by Raghavendra~\etal~in \cite{analytical1},
where a three stage approach combining pose and image texture components was proposed. 
Another method by Lijun~\etal~\cite{analytical2} combined several face-image characteristics, including alignment, occlusion and pose, into a pipeline for quality score calculation. Several conceptually similar approaches that exploit different (explicit) visual cues have been presented in the literature over the years~\cite{analy11, analy12, analy13, analy14}. However, the performance of such methods is typically 
limited, as they focus only on the characteristics of the input samples, with no regard to the utilized FR model. Nevertheless, a new group of analytical methods has recently emerged that incorporates information from both, the input face sample as well as the targeted FR system into the quality estimation process. An example of such an approach was presented by Terhörst~\etal~\cite{ser-fiq} in the form of the SER-FIQ technique. SER-FIQ calculates a quality score from the embedding variations of a given input face sample, caused by using different configurations of dropout layers. Another method, called FaceQAN by Babnik~\etal~\cite{faceqan}, relies on adversarial attacks (which are harder to generate for high quality images) to calculate quality scores. Both of these methods achieve excellent results, but are also comparably computationally demanding, due to their reliance on running several instances of the same sample through the given FR model.\vspace{-0.mm}

\noindent\textbf{Regression-based methods.} FIQA techniques from this group typically train a (quality) regression model using some sort of (pseudo) quality labels. Regression-based methods have received considerable attention over recent years, with most of the research exploring effective mechanisms for generating informative pseudo quality annotations. An early technique from this group, by Best-Rowden and Jain~\cite{bestrowden}, for example, used human raters to annotate the (perceived) quality of facial images, and then trained a quality predictor  on the resulting quality labels. Another technique, named FaceQnet~\cite{faceqnet1, faceqnet2}, relied on embedding comparisons with the highest quality image of each individual to estimate reference quality scores. Here, the highest quality images of each individual were determined using an external quality compliance tool and a ResNet-based regressor was then trained on the extracted quality labels. A more recent approach, called PCNet~\cite{pcnet}, used a large number of mated image pairs (i.e., a pair of distinct images of the same individual), to train a CNN-based regression model, where the quality labels were defined by the embedding similarity of the mated pairs. The SDD-FIQA approach, by Ou~\etal~\cite{sdd-fiqa}, extended this concept to also include non-mated (impostor) pairs, (i.e., two unique images of different individuals), where the label for a single image was computed as the Wasserstein distance between the mated and non-mated score distributions. LightQNet, by Chen~\etal~\cite{lightqnet}, trained a lightweight model, by employing an identification quality loss using quality scores computed from various image comparisons. While regression-based methods in general perform well over a variety of benchmarks and state-of-the-art FR models, their main weakness is the lack of specialization. As the optimal quality estimate for a given input image, is \textit{by definition} FR model specific~\cite{eval2,iwbf2023}, regression-based techniques may require retraining towards the targeted FR model to ensure ideal performance. \vspace{0.8mm}

\noindent\textbf{Model-based methods.} The last group of techniques combines face-image quality assessment and face recognition into a single task. One such technique, PFE by Shi and Jain~\cite{pfe}, learned to predict a pair of vectors from the input image, i.e., a mean and a variance vector. The mean vector can be thought of as the embedding of the input sample image, while the variance vector represents the sample's variability, and can be used to calculate the sample quality. The presented method inspired several new (uncertainty-aware) methods~\cite{pfe1, pfe2, pfe3}, further improving on the performance of PFE. Another notable technique, called MagFace~\cite{magface}, extended the popular ArcFace loss~\cite{arcface} by incorporating a magnitude-aware angular margin term, which dynamically adjusts class boundaries. The embeddings produced by MagFace encode quality in the magnitude of the embedding itself and can hence be easily inferred. A powerful FIQA technique, called the CR-FIQA, was recently proposed by Boutros~\etal in~\cite{cr-fiqa}. CR-FIQA calculates the quality of the input samples as the ratio between the positive class center and nearest negative class center in a classification task setting, and was demonstrated to produce highly competitive results across various datasets and  settings. \vspace{0.8mm} 

\noindent\textbf{Our contribution.} The DifFIQA technique, the \textit{main contribution} of this work, can be seen as an analytical method that relies on the capabilities of a DDPM in combination with a chosen FR model. From a conceptual point of view, it is most closely related to FaceQgen~\cite{faceqgen}, a FIQA technique that uses a (GAN-based) generator model for synthesizing high-quality versions of the input samples and the respective discriminator (that aims to distinguish between genuine high-quality images and poorly restored ones) for quality scoring. 
Unlike FaceQgen, which analyzes the differences between the original and restored images independently of the target FR model,  DifFIQA utilizes results from the forward (i.e., noising/degradation) as well as backward (denoising/restoration) diffusion processes and quantifies the embedding variability/uncertainty in the embedding space of a selected FR model for quality estimation. As we show later in the experimental section, this leads to highly competitive FIQA results when compared to the state-of-the-art. 

\section{Methodology}\label{sec:methodology}

The stability of the image representations in the embedding space of a given FR model is highly indicative of the input-image quality, as demonstrated by the success of various recent FIQA techniques \cite{ser-fiq,faceqan}. One way to explore this stability is by causing perturbations in the image space and analyzing the impact of the perturbations in the embedding space of the targeted FR model. This can, for example, be achieved by using the forward and backward processes of modern diffusion approaches where: the forward process adds some amount of noise to the sample, and the backward process tries to remove the noise, by reconstructing the original. 
Our main contribution, the DifFIQA technique, takes advantage of the proposed idea, as illustrated in Figure~\ref{fig:overview}, and employs a custom DDPM model for the generation of noisy and reconstructed images. The generated images are then passed through a chosen FR model to explore the impact of the perturbations on the variability of the embedding corresponding to the input image.

\subsection{Preliminaries}\label{sec:methodology:diffusion}


To make the paper self-contained, we briefly present the main concept behind denoising diffusion probabilistic models (DDPMs), with a focus on their application within our approach. More information on the theoretical background and applications of diffusion models can be found in~\cite{diff_survey1}. 

In general, DDPMs represent a special type of generative model that learns to model (image) data distributions through two types of processes: a forward (noising) process and backward (denoising) process \cite{iwbf2023ddpm,diff_survey1}. The \textbf{forward diffusion process} $\mathcal{F}_d$ iteratively adds noise to the given input image $x_0$, by sampling from a Gaussian distribution $\mathcal{N}(0,I)$. The result of this process is a noisy sample $x_t$, where $t$ is the number of time steps chosen from the sequence $\{0, 1, \dots,  T\}$. The whole forward process $\mathcal{F}_d$ can be presented as a Markov chain given by
\begin{equation}\label{eq:forward_markov_chain}
    q(x_t|x_{t-1}) = \mathcal{N}(x_t|x_{t-1}\sqrt{1 - \beta_t}, \beta_t I),
\end{equation}
where $\beta_t$ is a variance parameter  that defines how much noise is added to the sample at the time instance $t$ of the forward process. By making use of the reparameterization trick~\cite{forward_trick, forward_trick2}, any sample $x_t$ can be obtained directly from the input sample $x_0$, i.e.:
\begin{equation}\label{eq:forward_diffusion}
    q(x_t|x_0) = \mathcal{N}(x_t;\sqrt{\overline{\alpha}_t} x_0, (1-\overline{\alpha}_t)I),
\end{equation}
where $\overline{\alpha}_t = \prod^t_{i=0} (1 - \beta_i)$.

The \textbf{backward diffusion process} $\mathcal{B}_d$ attempts to iteratively denoise the generated samples $x_t$, using a deep neural network model ${D}_{\theta}$ parameterized by $\theta$, according to
\begin{equation}
    p(x_{t-1}|x_t) = \mathcal{N}(x_{t-1};\mu_\theta(x_t,t),\tilde{\beta}_tI),
\end{equation}
where $t = T, \dots, 0$, $\tilde{\beta}_t = \frac{1 - \overline{\alpha}_{t-1}}{1 - \overline{\alpha}_t}\beta_t$, and
\begin{equation}
\mu_\theta(x_t,t) = \frac{\sqrt{\overline{\alpha}_{t-1}}\beta_t}{1 - \overline{\alpha_t}}x_0 + \frac{\sqrt{\alpha_t}(1 - \overline{\alpha}_{t-1})}{1-\overline{\alpha}_t}x_t.
\end{equation}
The network is trained to optimize $\mu_\theta$, by minimizing the $\mathcal{L}_2$ loss
\begin{equation}\label{eq:backward_diffusion}
    \mathcal{L}_2 = \mathbb{E}_{t,x_0}||D_\theta(x_t, t) - x_0||^2,
\end{equation}
where $D_\theta(x_t, t)$ is the reconstructed and $x_0$ the input image. In the remainder of the paper, we drop the subscript $\theta$ and use $D$ to denote the deep neural network, which is represented by an unconditional UNet model. 

\subsection{Overview of DifFIQA}\label{sec:methodology:method_overview}

Given a face image $x$, the goal of DifFIQA is to estimate the quality score $q_x \in \mathbb{R}$, by exploring the effects of the forward and backward diffusion processes of a custom DDPM model $D$ on the image representation $e_x$ in the embedding space of a given FR model $M$. DifFIQA consists of two main steps, dedicated to: $(i)$ \textit{image perturbation} and $(ii)$ \textit{quality-score calculation}. 
The image perturbation step uses the forward diffusion process $\mathcal{F}_d$ to create a noisy sample $x_t$ from the input image $x$  and the backward process $\mathcal{B}_d$ to generate the restored (denoised) image $\hat{x}$. 
In the quality-score calculation step, the representations $e_x$, $e_{x_t}$, $e_{\hat{x}}$ corresponding to the input $x$, noisy $x_t$ and restored image $\hat{x}$, are calculated using the FR model $M$ and then analyzed for disparities to infer the final 
quality score $q_x$ of the input sample $x$. To also capture pose-related quality information, DifFIQA repeats the entire process using a horizontally flipped version $x^f$ of the input image $x$, as also illustrated in Figure~\ref{fig:overview}.    

\begin{figure}[!ht]
    \centering
    \includegraphics[width=0.99\columnwidth]{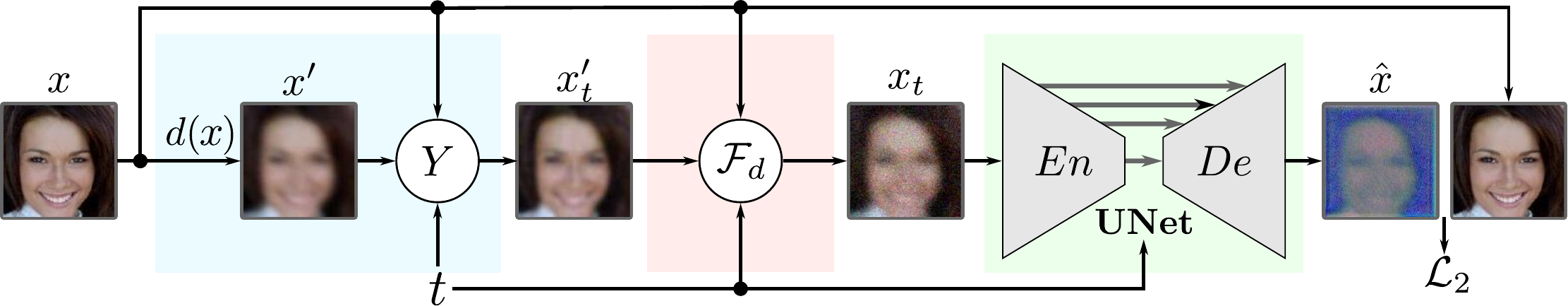}\vspace{1mm}
    \caption{\textbf{Presentation of the extended DDPM learning scheme.} Given a training sample $x$ and a time step $t$, the proposed approach generates a time step dependent degraded image $x'_t$, by combining the original with a degraded image using the function $Y$. The image $x'_t$ is then used to generate a noisy sample $x_t$ using the standard forward diffusion approach $\mathcal{F}_d$. A UNet ($D$) is then trained to reconstruct the input sample in the backward process $\mathcal{B}_d$.}\vspace{-2mm}
    \label{fig:training}
\end{figure}

\subsection{Extended DDPM Training}\label{sec:methodology:trainig_process}

To train the DDPM model $D$ (i.e., a UNet \cite{ronneberger2015u}) needed by DifFIQA, we extend the standard training process of diffusion models to incorporate time dependent image degradations, as illustrated in Figure \ref{fig:training}. These additional (time-dependent) degradations allow the model to learn to gradually reverse these degradations and, in turn, to construct higher quality images during the backward process. 
Formally, given an input face image $x_0 = x$, the training procedure first constructs a degraded image $x' = d(x)$, where $d(\cdot)$ is some degradation function. A time step $t \in [0, T]$ is then selected for the given sample, from which a time-dependent degraded image is computed as follows:
\begin{equation}\label{eq:time_dependant_degradation}
    x'_t = Y(x_0, x', t) = (1 - \ddot{\alpha}_t) x_0 + \ddot{\alpha}_t x'
\end{equation}
where $\ddot{\alpha}_t$ is calculated as $\sin(\frac{t}{T}\cdot\frac{\pi}{2})$. The degraded image $x'_t$ is then used to produce the noisy sample $x_t$ using \eqref{eq:forward_diffusion}. 
Here,  $\ddot{\alpha}_t$ is a time-dependent variable that monotonically increases on the interval $t \in [0, T]$, such that $\ddot{\alpha}_{t=0} = 0$ and $\ddot{\alpha}_{t=T} = 1$. In other words, at time step $0$ only the non-degraded image is considered, while at time step $T$ only the degraded image is considered. To implement the degradation function  $d(\cdot)$, we use  part of the BSRGAN~\cite{bsrgan} framework that creates a random sequence of image mappings that imitate real-life degradations. 

Diffusion models are commonly trained on the full range of time steps $[1, T]$ and learn to generate images from pure noise. However, such a setting is not relevant in the context of quality assessments, as the generated (denoised) images have to exhibit a sufficient correspondence with  the input samples $x$. The easiest solution to this issue is to limit the number of time steps, on which the model is trained $t \in [1, T']$, where $T' < T$, and, in turn, ensure that the noisy image is properly conditioned on the input $x$. The extended training procedure then minimizes \eqref{eq:backward_diffusion} until convergence.





\subsection{Generating Noisy and Reconstructed Images}\label{sec:methodology:generating_images}


To estimate the quality of a given face image $x$, DifFIQA makes use of the forward and backward diffusion processes of the trained DDPM. 
Because head pose is an important factor of face quality, which the underlying DDPM can not explicitly account for, 
we extend our methodology, by first constructing a horizontally flipped image $x^f$ that we utilize alongside the original image $x$ in the quality-score calculation step, similarly to~\cite{faceqan}. The main intuition behind this approach is to exploit the symmetry of human faces, where large deviations from frontal pose induce large disparities between the embeddings of the original and flipped images that can be quantified during quality estimation.  Thus, for a pair of input face images $(x, x^f)$ and a given time step $t$, we construct a pair of noisy 
 $(x_t, x_t^f)$ and restored images $(\hat{x}, \hat{x}^f)$ and use the generated data for quality estimation.

\subsection{Quality-Score Calculation}\label{sec:methodology:quality_extraction}

DifFIQA relies on the assumption that the embeddings of lower-quality images are more sensitive to image perturbations introduced by the forward and backward diffusion processes than higher-quality images. To quantify this sensitivity, we calculate the average cosine similarity between the embedding of the input image $x$ and all generated noisy and restored counterparts. Additionally, since diffusion models rely on the (random) sampling from a normal distribution, we repeat the whole process $n$ times and average the results, i.e.,\vspace{-2mm}
\begin{equation}
    q_x = \frac{1}{n|\mathcal{E}|} \sum^n_{i=1}\sum_{e_y\in\mathcal{E}} \frac{e_x^T e_y}{\|e_x\| \cdot \|e_y\|},  
\end{equation}
where $\mathcal{E}$ is a set of generated image embeddings, i.e,
    $\mathcal{E} = \{ e_{x_t}, e_{\hat{x}}, e_{x^f}, e_{x^f_t}, e_{\hat{x}^f}\}$,
computed with the FR model $M$ as $e_z = M(x_z)$. In the above equation, the operator $|\cdot|$ denotes the set cardinality and $\|\cdot\|$ the $L_2$ norm.

\begin{table}[!t!]
    \centering
    \renewcommand{\arraystretch}{1.1}
    \caption{\textbf{Summary of the characteristics of the experimental datasets.} We evaluate DifFIQA across seven diverse datasets with different quality factors and of different size.\vspace{1mm}}
\resizebox{\columnwidth}{!}{%
    \begin{tabular}{llcccccccc}
        \toprule
        \multirow{ 2}{*}{\textbf{Dataset}} & \multirow{ 2}{*}{\textbf{\#Images}} & \multirow{ 2}{*}{\textbf{\#IDs}} & \multicolumn{2}{c}{\textbf{\#Comparisons}} && \multicolumn{4}{c}{\textbf{Main Quality Factors}$^\dagger$$^\ddagger$}\\\cline{4-5}\cline{7-10}
        
        & & & Mated & Non-mated && Pose & O-E & B-R-N & Sc\\ 
        \midrule
        Adience~\cite{adience} & $19{\small,}370$ & $2{\small,}284$ & $20{\small,}000$ & $20{\small,}000$ && M & M & L & M \\
        CALFW~\cite{calfw} & $12{\small,}174$ & $4{\small,}025$ & $3{\small,}000$ & $3{\small,}000$ && M & M & L & M \\
        CFP-FP~\cite{cfp-fp} & $7{\small,}000$ & $500$ & $3{\small,}500$ & $3{\small,}500$ && H & L & L & M  \\
        CPLFW~\cite{cplfw} & $11{\small,}652$ & $3{\small,}930$& $3{\small,}000$ & $3{\small,}000$ && H & L & M & M \\
        IJB-C~\cite{ijbc} & $23{\small,}124^{\dagger\dagger}$ & $3{\small,}531$ & $19{\small,}557$ & $15{\small,}638{\small,}932$ && H & H & H & Lr\\
        LFW~\cite{lfw} & $13{\small,}233$ & $5{\small,}749$ & $3{\small,}000$ & $3{\small,}000$ && L & L & L & M\\
        XQLFW~\cite{xqlfw} & $13{\small,}233$ & $5{\small,}749$ & $3{\small,}000$ & $3{\small,}000$ && L & L & H & M\\
        \bottomrule
        \multicolumn{10}{l}{\footnotesize $^\dagger$O-E - Occlusion, Expression; B-R-N - Blur, Resolution, Noise; Sc - Scale.}\\
        \multicolumn{8}{l}{\footnotesize $^\ddagger$L - Low; M - Medium; H - High; Lr - Large; Values estimated subjectively by the authors.}\\
        \multicolumn{8}{l}{\footnotesize $^{\dagger\dagger}$ number of templates, each containing several images}
        \vspace{-5mm}
    \end{tabular}
    }
    \label{tab:my_flips}
\end{table}

\subsection{Model Distillation}\label{sec:methodology:quality_regression_model}

One of the main shortcomings of DifFIQA (and diffusion models in general) is the high computational complexity compared to other types of FIQA techniques. This complexity stems from the iterative nature of the backward diffusion process, which requires a large number of forward passes through the generative network. Since our approach repeats this process $n$-times, this only exacerbates the problem and adversely affects the applicability of the technique in real-world applications. To address this problem, we \textbf{distill the knowledge} encoded by DifFIQA into a regression model. Specifically, we select a pretrained CosFace FR model augmented with a (quality) regression head and fine-tune it on roughly two million quality labels extracted from the VGGFace2~\cite{vggface2} dataset using the proposed DifFIQA technique. Here, the labels are  normalized to $[0, 1]$ and then split into train and validation sets for the training procedure.
We refer to the distilled CosFace model as DifFIQA(R) hereafter, and evaluate it together with the original DifFIQA technique in the following sections. 



\begin{figure*}[t]
    \centering
    \includegraphics[width=0.85\textwidth]{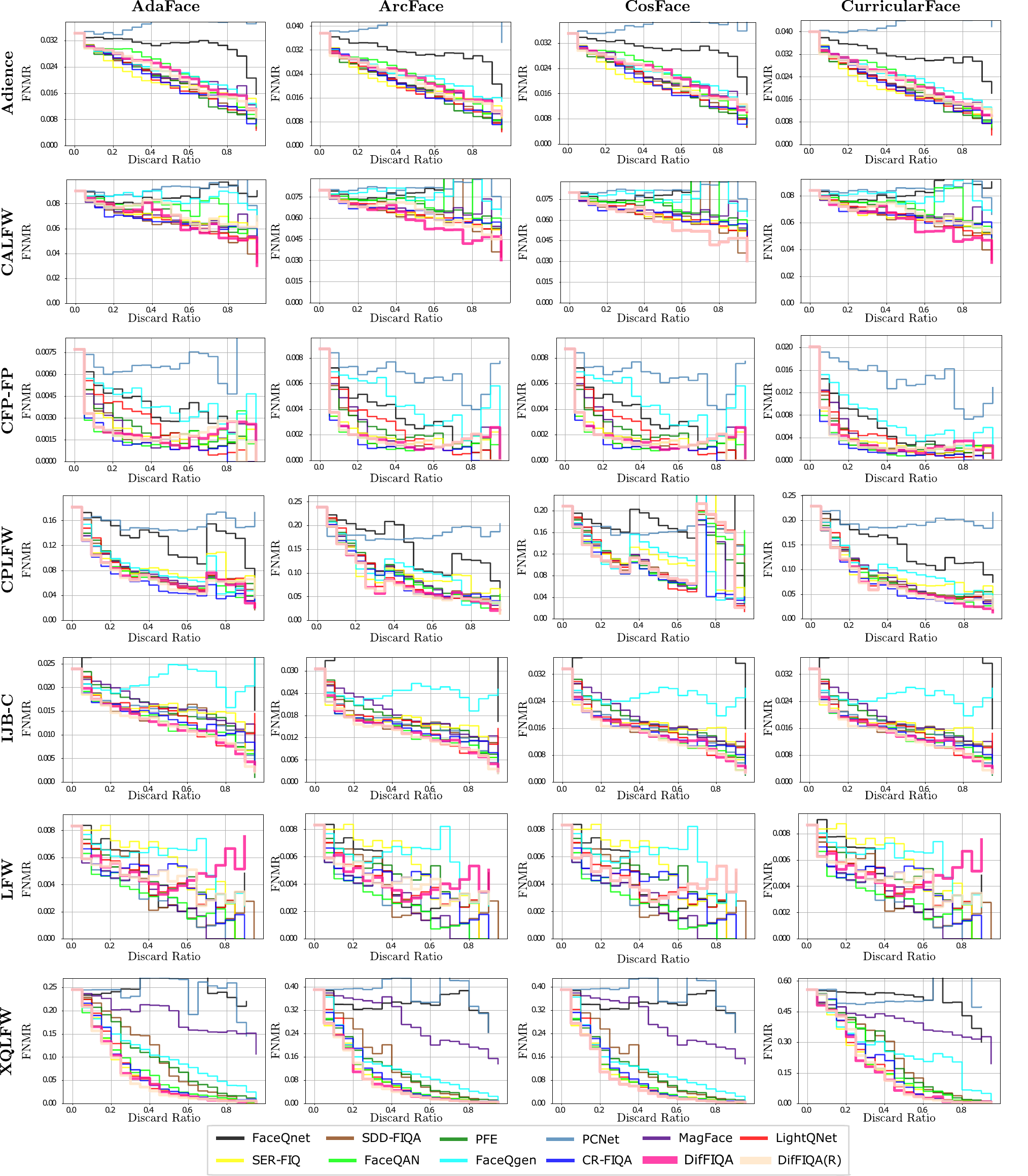}\vspace{1mm}
    \caption{\textbf{Comparison to the state-of-the-art in the form of (non-interpolated) EDC curves.} Results are presented for seven diverse datasets, four FR models and in comparison to ten recent FIQA competitors. Observe how the distilled model performs comparably to the non-distilled version, especially at low discard rates. DifFIQA and DifFIQA(R) most convincingly outperform the competitors on the most chalenging IJB-C and XQLFW datasets. The figure is best viewed in color.}\vspace{-3mm}
    \label{fig:edc_curves}
\end{figure*}

\section{Experiments and Results}\label{sec:experiments_results}

\subsection{Experimental Setup}\label{sec:experiments_results:setting}

\noindent\textbf{Experimental setting.} We analyze the performance of Face\-QDiff in comparison to $10$ state-of-the-art FIQA methods, i.e.: $(i)$  the \textbf{analytical} FaceQAN~\cite{faceqan}, SER-FIQ~\cite{ser-fiq}, and FaceQgen \cite{faceqgen} models, $(ii)$ the \textbf{regression-based}  FaceQnet~\cite{faceqnet}, SDD-FIQA \cite{sdd-fiqa},  PCNet \cite{pcnet}, and LightQnet \cite{lightqnet} techniques, and $(iii)$ the \textbf{model-based}  MagFace \cite{magface}, PFE \cite{pfe}, and CR-FIQA \cite{cr-fiqa} methods. We test all methods on $7$ commonly used benchmarks with different quality charcteristics, as summarized in Table~\ref{tab:my_flips}, i.e.: Adience~\cite{adience}, Cross-Age Labeled Faces in the Wild (CALFW) \cite{calfw}, Celebrities in Frontal-Profile in the Wild (CFP-FP) \cite{cfp-fp}, Cross-Pose Labeled Faces in the Wild (CPLFW) \cite{cplfw}, large-scale IARPA Janus Benchmark C~(IJB-C) \cite{ijbc}, Labeled Faces in the Wild (LFW) \cite{lfw} and the Cross-Quality Labeled Faces in the Wild~(XQLFW) \cite{xqlfw}. Because the performance of FIQA techniques is dependent on the FR model used, we investigate how well the techniques generalize over $4$ state-of-the-art models, i.e.: AdaFace\footnote{\label{fnote:adaface}\scriptsize{\url{https://github.com/mk-minchul/AdaFace}}} \cite{adaface}, ArcFace\footnote{\label{fnote:arcface}\scriptsize{\url{https://github.com/deepinsight/insightface}}} \cite{arcface}, CosFace\footref{fnote:arcface} \cite{cosface}, and CurricularFace\footnote{\label{fnote:curricularface}\scriptsize{\url{https://github.com/HuangYG123/CurricularFace}}} \cite{curricularface} - all named after their training losses. All FR models use a ResNet100 backbone, and are trained on the WebFace12M\footref{fnote:adaface}, MS1MV3\footref{fnote:arcface}, Glint360k\footref{fnote:arcface}, and CASIA-WebFace\footref{fnote:curricularface} datasets.\vspace{0.8mm}

\noindent\textbf{Evaluation methodology.} Following standard evaluation methology \cite{faceqan,ser-fiq,cr-fiqa} and taking recent insights into account \cite{eval1,eval2}, we evaluate the performance of DifFIQA using \textit{non-interpolated} Error-versus-Discard Characteristic (EDC) curves (often also referred to as Error-versus-Reject Characteristic or ERC curves in the literature) and the consequent pAUC~(partial Area Under the Curve) values. The EDC curves measure the False Non-Match Rate~(FNMR), given a predefined False Match Rate~(FMR) ($10^{-3}$ in our case), with increasing low-quality image discard (reject) rates. In other words, EDC curves measure how the performance of a given FR model improves when some percentage of the lowest quality images is discarded. Since rejecting a large percentage of all samples is not feasible/practical in real-world application scenarios, we are typically most interested in the performance at the lower discard rates. For this reason we report the pAUC values, where only the results up to a predetermined drop rate threshold are considered. Furthermore, for easier interpretation and comparison of scores over different dataset, we normalize the calculated pAUC values using the FNMR at $0\%$ discard rate
, with lower pAUC values indicating better performance.\vspace{0.8mm}

\noindent\textbf{Implementation Details.} During training of the DDPM, the maximum number of forward steps is set to $T=1000$, yet the underlying model is trained only using up to $T'=100$ forward diffusion steps. 
The value of $T'$ does not define the number of time steps $t$ taken at inference time, it only sets the possible upper bound. 
This process ensures that images produced by the forward process are only partially noisy, so the backward process is properly conditioned on the input image and learns to restore it during training. To account for the randomness introduced by the forward process, we repeat the diffusion process $n=10$ times and average the results over all iterations, when computing the final quality score. The utilized UNet model ($D$) consists of four downsampling and upsampling modules, each decreasing (increasing) the dimensions of the representations by a factor of two. Training is done using the Adam optimizer, with a learning rate of $8.0e^{-5}$ in combination with an Exponential Moving Average~(EMA) model, with a decay rate of $0.995$. The presented hyperparameters were determined through preliminary experiments on hold-out data to ensure a reasonable trade-off between training speed and reproducible performance. All experiments were conducted on a desktop PC with an Intel i9-10900KF CPU, $64$ GB of RAM and an Nvidia $3090$ GPU.

\subsection{Comparison with the State-of-the-Art}\label{sec:experiments_results:results}

In this section, we compare DifFIQA and the distilled version, DifFIQA(R), with ten state-of-the-art competitors and analyze: $(i)$ the \textit{performance characteristics} of the tested techniques, and $(ii)$ their \textit{runtime complexity}.\vspace{0.8mm}

\begin{table}[t]
    \centering
    \caption{\textbf{Comparison to the state-of-the-art.} The table reports pAUC scores at a discard rate of $0.3$ and a FMR of $10^{-3}$. Average results across all datasets are marked $\overline{\text{pAUC}}$. The best result for each dataset is shown in bold, the overall best result is colored green, the second best blue and the third best red. \vspace{.5mm}}
    \resizebox{\columnwidth}{!}{%
    \begin{tabular}{l | l l l l l l l | r}

\toprule
\multicolumn{9}{c}{\textbf{AdaFace - pAUC@FMR=}$10^{-3}$ ($\downarrow$)} \\
\midrule
\textbf{FIQA model} & \textbf{Adience} & \textbf{CALFW} & \textbf{CFP-FP} & \textbf{CPLFW} & \textbf{IJB-C} & \textbf{LFW} & \textbf{XQLFW} & $\overline{\text{pAUC}}$ \\ 
\cmidrule{1-9}
{FaceQnet} \cite{faceqnet}&  $0.963$ &  $0.938$ &  $0.717$ &  $0.887$ &  $1.256$ &  $0.884$ &  $0.977$ &  $0.946$ \\
{SDD-FIQA} \cite{sdd-fiqa}&  $0.839$ &  $0.871$ &  $0.500$ &  $0.688$ &  $0.782$ &  $0.825$ &  $0.842$ &  $0.764$ \\
{PFE} \cite{pfe}&  $0.833$ &  $0.890$ &  $0.566$ &  $0.681$ &  $0.868$ &  $0.771$ &  $0.798$ &  $0.772$ \\
{PCNet} \cite{pcnet}&  $1.005$ &  $0.979$ &  $0.862$ &  $0.898$ &  $0.788$ &  $0.661$ &  $0.987$ &  $0.883$ \\
{MagFace} \cite{magface}&  $0.860$ &  $0.866$ &  $0.524$ &  $0.664$ &  $0.883$ &  $0.666$ &  $0.913$ &  $0.768$ \\
{LightQNet} \cite{lightqnet}&  $0.847$ &  $0.894$ &  $0.641$ &  $0.684$ &  $0.797$ &  $0.777$ &  $0.704$ &  $0.763$ \\
{SER-FIQ}$^{\dagger}$ \cite{serfiq}&  $\mathbf{0.807}$ &  $0.892$ &  $0.475$ &  $0.626$ &  $0.762$ &  $0.935$ &  $n/a$ &  $0.749$ \\
{FaceQAN} \cite{faceqan}&  $0.890$ &  $0.919$ &  $\mathbf{0.383}$ &  $0.619$ &  $0.756$ &  $\mathbf{0.656}$ &  $0.654$ &  \cellcolor{red!10}$\mathbf{0.697}$ \\
{CR-FIQA} \cite{cr-fiqa} &  $0.844$ &  $\mathbf{0.851}$ &  $0.391$ &  $\mathbf{0.588}$ &  $0.750$ &  $0.707$ &  $0.684$ &  \cellcolor{green!10}$\mathbf{0.688}$ \\
{FaceQgen} \cite{faceqgen}&  $0.858$ &  $0.970$ &  $0.718$ &  $0.694$ &  $0.853$ &  $0.834$ &  $0.736$ &  $0.809$ \\
\midrule
{DifFIQA} (ours)&  $0.864$ &  $0.900$ &  $0.416$ &  $0.608$ &  $0.761$ &  $0.719$ &  $0.627$ &  $0.699$ \\
{DifFIQA(R)} (ours)&  $0.865$ &  $0.895$ &  $0.412$ &  $0.601$ &  $\mathbf{0.731}$ &  $0.708$ &  $\mathbf{0.610}$ &  \cellcolor{blue!10}$\mathbf{0.689}$ \\

\toprule
\multicolumn{9}{c}{\textbf{ArcFace - pAUC@FMR}=$10^{-3}$ ($\downarrow$)} \\
\midrule
\textbf{FIQA model} & \textbf{Adience} & \textbf{CALFW} & \textbf{CFP-FP} & \textbf{CPLFW} & \textbf{IJB-C} & \textbf{LFW} & \textbf{XQLFW} & $\overline{\text{pAUC}}$ \\ 
\cmidrule{1-9}
{FaceQnet} \cite{faceqnet} &  $0.943$ &  $0.955$ &  $0.693$ &  $0.878$ &  $1.224$ &  $0.884$ &  $0.899$ &  $0.925$ \\
{SDD-FIQA} \cite{sdd-fiqa}&  $0.783$ &  $0.901$ &  $0.491$ &  $0.734$ &  $0.720$ &  $0.808$ &  $0.774$ &  $0.744$ \\
{PFE} \cite{pfe}&  $0.774$ &  $0.932$ &  $0.524$ &  $0.738$ &  $0.783$ &  $0.779$ &  $0.641$ &  $0.739$ \\
{PCNet} \cite{pcnet}&  $1.022$ &  $1.006$ &  $0.868$ &  $0.783$ &  $0.706$ &  $\mathbf{0.623}$ &  $1.004$ &  $0.859$ \\
{MagFace} \cite{magface}&  $0.812$ &  $0.902$ &  $0.549$ &  $0.717$ &  $0.824$ &  $0.635$ &  $0.943$ &  $0.769$ \\
{LightQNet} \cite{lightqnet}&  $0.789$ &  $0.913$ &  $0.612$ &  $0.752$ &  $0.721$ &  $0.745$ &  $0.621$ &  $0.736$ \\
{SER-FIQ}$^{\dagger}$ \cite{ser-fiq}&  $\mathbf{0.767}$ &  $0.903$ &  $0.416$ &  $0.656$ &  $0.671$ &  $0.935$ &  $n/a$ &  $0.724$ \\
{FaceQAN} \cite{faceqan}&  $0.824$ &  $0.941$ &  $0.373$ &  $0.677$ &  $0.673$ &  $0.624$ &  $0.581$ &  \cellcolor{red!10}$\mathbf{0.670}$ \\
{CR-FIQA} \cite{cr-fiqa}&  $0.808$ &  $\mathbf{0.891}$ &  $\mathbf{0.358}$ &  $0.689$ &  $0.664$ &  $0.675$ &  $0.642$ &  $0.675$ \\
{FaceQgen} \cite{faceqgen} &  $0.817$ &  $0.985$ &  $0.784$ &  $0.701$ &  $0.785$ &  $0.802$ &  $0.653$ &  $0.789$ \\
\midrule
{DifFIQA} (ours)&  $0.805$ &  $0.900$ &  $0.399$ &  $0.647$ &  $0.675$ &  $0.695$ &  $\mathbf{0.546}$ &  \cellcolor{blue!10}$\mathbf{0.667}$ \\
{DifFIQA(R)} (ours)&  $0.801$ &  $0.898$ &  $0.389$ &  $\mathbf{0.646}$ &  $\mathbf{0.655}$ &  $0.708$ &  $0.554$ &  \cellcolor{green!10}$\mathbf{0.665}$ \\

\toprule
\multicolumn{9}{c}{\textbf{CosFace - pAUC@FMR=}$10^{-3}$ ($\downarrow$)} \\
\midrule
\textbf{FIQA model} & \textbf{Adience} & \textbf{CALFW} & \textbf{CFP-FP} & \textbf{CPLFW} & \textbf{IJB-C} & \textbf{LFW} & \textbf{XQLFW} & $\overline{\text{pAUC}}$ \\ 
\cmidrule{1-9}
{FaceQnet} \cite{faceqnet}&  $0.952$ &  $0.955$ &  $0.693$ &  $0.879$ &  $1.248$ &  $0.884$ &  $0.899$ &  $0.930$ \\
{SDD-FIQA} \cite{sdd-fiqa}&  $0.825$ &  $0.901$ &  $0.491$ &  $0.735$ &  $0.721$ &  $0.808$ &  $0.774$ &  $0.751$ \\
{PFE} \cite{pfe}&  $0.813$ &  $0.932$ &  $0.524$ &  $0.748$ &  $0.784$ &  $0.779$ &  $0.641$ &  $0.746$ \\
{PCNet} \cite{pcnet}&  $1.009$ &  $1.006$ &  $0.868$ &  $0.835$ &  $0.710$ &  $\mathbf{0.623}$ &  $1.004$ &  $0.865$ \\
{MagFace} \cite{magface}&  $0.852$ &  $0.902$ &  $0.549$ &  $0.724$ &  $0.821$ &  $0.635$ &  $0.943$ &  $0.775$ \\
{LightQNet} \cite{lightqnet}&  $0.835$ &  $0.913$ &  $0.612$ &  $0.753$ &  $0.713$ &  $0.745$ &  $0.621$ &  $0.742$ \\
{SER-FIQ}$^{\dagger}$ \cite{ser-fiq}&  $\mathbf{0.793}$ &  $0.903$ &  $0.416$ &  $0.711$ &  $0.661$ &  $0.935$ &  $n/a$ &  $0.736$ \\
{FaceQAN} \cite{faceqan}&  $0.871$ &  $0.941$ &  $0.373$ &  $\mathbf{0.667}$ &  $0.675$ &  $0.624$ &  $0.581$ &  \cellcolor{red!10}$\mathbf{0.676}$ \\
{CR-FIQA} \cite{cr-fiqa}&  $0.835$ &  $\mathbf{0.891}$ &  $\mathbf{0.358}$ &  $0.681$ &  $0.664$ &  $0.675$ &  $0.642$ &  $0.678$ \\
{FaceQgen} \cite{faceqgen}&  $0.847$ &  $0.985$ &  $0.784$ &  $0.702$ &  $0.783$ &  $0.802$ &  $0.653$ &  $0.794$ \\
\midrule
{DifFIQA} (ours)&  $0.841$ &  $0.900$ &  $0.399$ &  $0.669$ &  $0.672$ &  $0.695$ &  $\mathbf{0.546}$ &  \cellcolor{blue!10}$\mathbf{0.675}$ \\
{DifFIQA(R)} (ours)&  $0.838$ &  $0.900$ &  $0.389$ &  $0.669$ &  $\mathbf{0.644}$ &  $0.695$ &  $\mathbf{0.546}$ &  \cellcolor{green!10}$\mathbf{0.669}$ \\

\toprule
\multicolumn{9}{c}{\textbf{CurricularFace - pAUC@FMR=}$10^{-3}$ ($\downarrow$)} \\
\midrule
\textbf{FIQA model} & \textbf{Adience} & \textbf{CALFW} & \textbf{CFP-FP} & \textbf{CPLFW} & \textbf{IJB-C} & \textbf{LFW} & \textbf{XQLFW} & $\overline{\text{pAUC}}$ \\ 
\cmidrule{1-9}
{FaceQnet} \cite{faceqnet}&  $0.921$ &  $0.947$ &  $0.601$ &  $0.867$ &  $1.248$ &  $0.908$ &  $0.984$ &  $0.925$ \\
{SDD-FIQA} \cite{sdd-fiqa}&  $0.776$ &  $0.900$ &  $0.409$ &  $0.696$ &  $0.721$ &  $0.821$ &  $0.817$ &  $0.734$ \\
{PFE} \cite{pfe}&  $0.759$ &  $0.923$ &  $0.415$ &  $0.691$ &  $0.784$ &  $0.785$ &  $0.835$ &  $0.742$ \\
{PCNet} \cite{pcnet}&  $1.004$ &  $0.996$ &  $0.887$ &  $0.899$ &  $0.710$ &  $0.656$ &  $0.938$ &  $0.870$ \\
{MagFace} \cite{magface}&  $0.793$ &  $0.892$ &  $0.477$ &  $0.689$ &  $0.821$ &  $0.661$ &  $0.862$ &  $0.742$ \\
{LightQNet} \cite{lightqnet}&  $0.769$ &  $0.910$ &  $0.462$ &  $0.704$ &  $0.713$ &  $0.767$ &  $0.739$ &  $0.723$ \\
{SER-FIQ}$^{\dagger}$ \cite{ser-fiq}&  $\mathbf{0.750}$ &  $0.883$ &  $0.389$ &  $0.625$ &  $0.661$ &  $0.942$ &  $n/a$ &  $0.708$ \\
{FaceQAN} \cite{faceqan}&  $0.811$ &  $0.931$ &  $0.343$ &  $0.637$ &  $0.669$ &  $\mathbf{0.644}$ &  $0.835$ &  $0.696$ \\
{CR-FIQA} \cite{cr-fiqa}&  $0.797$ &  $\mathbf{0.877}$ &  $\mathbf{0.318}$ &  $\mathbf{0.615}$ &  $0.664$ &  $0.693$ &  $0.789$ &  \cellcolor{green!10}$\mathbf{0.679}$ \\
{FaceQgen} \cite{faceqgen}&  $0.815$ &  $0.974$ &  $0.662$ &  $0.698$ &  $0.783$ &  $0.845$ &  $0.750$ &  $0.790$ \\
\midrule
{DifFIQA} (ours)&  $0.806$ &  $0.884$ &  $0.384$ &  $0.624$ &  $0.672$ &  $0.711$ &  $\mathbf{0.736}$ &  \cellcolor{red!10}$\mathbf{0.688}$ \\
{DifFIQA(R)} (ours)&  $0.788$ &  $0.892$ &  $0.358$ &  $0.622$ &  $\mathbf{0.644}$ &  $0.724$ &  $0.768$ &  \cellcolor{blue!10}$\mathbf{0.685}$ \\

\bottomrule
\multicolumn{9}{l}{$^\dagger$SER-FIQ was used to create XQLFW, so the results here are not reported for a fair comparison.}\vspace{-7mm}
    \end{tabular}
    }
    \label{tab:pauc_values}
\end{table}

\begin{table*}[t]
    \centering
    \caption{\textbf{Runtime complexity.} The reported results (in ms) were computed over the XWLFW dataset and the same experimental hardware. Note how the destillation process leads to a speed-up of more than three orders of magnitude from DifFIQA to  DifFIQA(R). \vspace{1mm}}
    \resizebox{\textwidth}{!}{%
    \begin{tabular}{ l  cccccccccccc}
        \toprule
       \multirow{ 2}{*}{\textbf{FIQA model}} & \multirow{ 2}{*}{\textbf{CR-FIQA} \cite{cr-fiqa}} &\multirow{ 2}{*}{\textbf{SDD-FIQA} \cite{sdd-fiqa}} & \multirow{2}{*}{\textbf{FaceQAN} \cite{faceqan}} & \multirow{ 2}{*}{\textbf{MagFace} \cite{magface}} & \multirow{2}{*}{\textbf{SER-FIQ} \cite{ser-fiq}} & \multirow{ 2}{*}{\textbf{FaceQnet} \cite{faceqnet}} & \multirow{2}{*}{\textbf{FaceQgen} \cite{faceqgen}} & \multirow{ 2}{*}{\textbf{LightQnet} \cite{lightqnet}} & \multirow{2}{*}{\textbf{PCNet} \cite{pcnet}} & \multirow{ 2}{*}{\textbf{PFE} \cite{pfe}} & \multicolumn{2}{c}{\textbf{Ours}}\\ \cmidrule{12-13}
       &&&&&&&&&&& \textbf{DifFIQA} & \textbf{DifFIQA(R)}\\
        \midrule
         Runtime ($\mu \pm\sigma$) & $0.15\pm0.37$ & $0.62\pm0.36$ & $334.13\pm118.79$ &  $1.08\pm0.36$ & $112.93\pm33.81$ & $42.11\pm2.14$ & $42.11\pm2.05$ & $18.54\pm18.68$ & $17.06\pm0.34$ & $42.69\pm12.26$ &  $1074.62\pm11.45$ & $1.24\pm0.36$ \\
        \bottomrule
    \end{tabular}
    }\vspace{-3mm}
    \label{tab:time_complexity}
\end{table*}

\noindent \textbf{Performance analysis.} In Figure~\ref{fig:edc_curves}, we show the (non-interpolated) EDC curves for all tested FR models and datasets, and report the corresponding pAUC scores in Table~\ref{tab:pauc_values}. Following the suggestions in~\cite{eval1,eval2}, we chose a discard rate of $0.3$, when calculating the pAUC values, but also report additional results in the supplementary material. We observe that the proposed diffusion-based FIQA techniques result in highly competitive performance across all datasets and FR models. The distilled DifFIQA(R) model, for example, leads to the lowest average  $\overline{\text{pAUC}}$ score with the ArcFace and CosFace FR models, and is the runner-up with the AdaFace and CurricularFace models with $\overline{\text{pAUC}}$ scores comparable to the top performer CR-FIQA\footnote{In the supplementary material we show that with a discard rate of $0.2$, DifFIQA(R) is the top performer with $3$ of the $4$ FR models.}. Several interesting findings can be made from the reported results, e.g.: $(i)$ While the performance of DifFIQA and DifFIQA(R) is in general close, the distilled version has a slight edge over the original, which suggests that the distillation process infuses some additional information into the FIQA procedure through the FR-based regression model; $(ii)$ The proposed FIQA models are particularly competitive on the difficult large-scale IJB-C dataset, where the DifFIQA(R) approach consistently outperforms all competing baseline models. A similar observation can also be made for the challenging XQLFW dataset, where the diffusion-based models are again the top performers, which speaks of the  effectiveness of diffusion-based quality estimation. \vspace{0.8mm}

\noindent \textbf{Runtime complexity.} In Table~\ref{tab:time_complexity}, we compare the runtime complexity of the evaluated FIQA techniques (in ms). To ensure a fair comparison, we utilize $(i)$ the same experimental hardware for all methods  (described in Section~\ref{sec:experiments_results:setting}), $(ii)$ use the official code, released by the authors for all techniques, and $(iii)$ compute average runtimes and standard deviations over the entire XQLFW dataset. As can be seen, the original approach, DifFIQA, despite being highly competitive in terms of performance, is among the most computationally demanding due to the use of the complex diffusion processes. With around $1s$ on average per image, the runtime complexity of the model is even significantly higher than that of the FaceQAN or SER-FIQ techniques that require multiple passes through their networks to estimate quality and which are already among the slower FIQA models. However, the distillation process, allows to reduce the runtime by roughly 
three orders of magnitude (or by $99,9\%$), making the distilled DifFIQA(R) comparable to the faster models evaluated in this experiment.

\begin{table}[t]
    \centering
    \caption{\textbf{Results of the ablation study.} The results are reported in terms of pAUC ($\downarrow$) at a FMR of $10^{-3}$ and a discard rate of $0.3$. \vspace{1mm}}
    \resizebox{\columnwidth}{!}{%
    \begin{tabular}{l | rrrr|r}
    \toprule
      \textbf{Model variant}  & \textbf{LFW} & \textbf{CPLFW} & \textbf{CALFW} & \textbf{XQLFW} & ${\overline{\text{pAUC}}}$  \\
    \midrule
    (A1): w/o Image Flipping & $0.702$ & $0.727$ & $0.888$ & $0.535$ & $0.713$ \\
    (A2): w/o Forward Pass & $0.730$ & $0.684$ & $0.897$ & $0.531$ & $0.710$\\
    (A3): DifFIQA $(t=20)$ & $0.657$ & $0.694$ & $0.945$ & $0.628$ & $0.731$\\
    \midrule
    DifFIQA (complete) & $0.695$ & $0.669$ & $0.900$ & $0.546$ & $\mathbf{0.702}$\\   
    \bottomrule         
    \end{tabular}
    }
    \label{tab:ablation_study}
\end{table}

\begin{figure}
    \centering
    \includegraphics[width=0.95\linewidth]{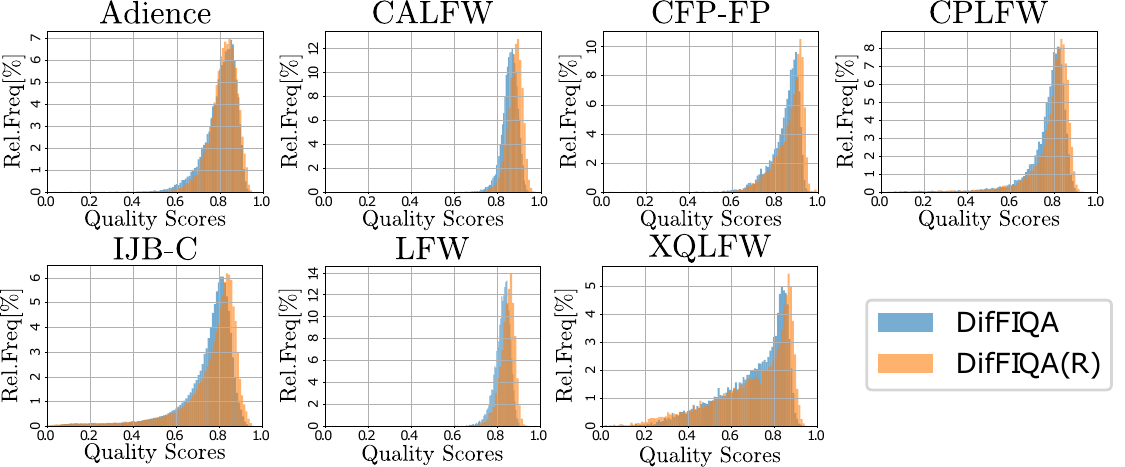}
    \caption{\textbf{Quality-score distributions.} DifFIQA  by DifFIQA(R) produce very consistent distributions over all seven test datasets.}\vspace{-3mm}
    \label{fig:distributions}
\end{figure}

\subsection{Ablation Study}\label{sec:experiments_results:ablation}

We perform several ablation studies to explore the impact of the main components of DifFIQA. Specifically, we are interested in: (A1) the impact of the flipping procedure, utilized to capture pose-related quality factors, (A2) the contribution of the forward pass (i.e., the noising step of the diffusion), and (A3) the impact of the number of forward diffusion steps $t$, where a larger number corresponds to higher amounts of noise in the image $x_t$ produced by the forward process. Because the ablations are only relevant for the (non-distilled) approach, we experiment solely with the DifFIQA technique and report results using the CosFace FR model and four datasets that feature  a broad range of quality characteritics, i.e., LFW, CPLFW, CALFW and XQLFW.   

From the results in Table~\ref{tab:ablation_study}, we observe that the exclusion of the flipping operation significantly degrades performance on the cross-pose (CPLFW) dataset, while contributing to minor improvements on CALFW and XQLFW. However, given that pose is considered one of the main factors still adversely affecting modern FR models, the flipping operation still helps with the performance across all the test datasets (see average A1 results). When removing the forward pass (in A2), we again see considerable performance drops on LFW and CPLFW, leading to lower average pAUC scores. This suggest that both (forward and backward) processes are important for good results across different datasets. Finally, we see that lesser amounts of noise and stronger conditioning on the input images leads to better results as evidenced by the A3 results with our model with 20 timesteps, instead of the 5 utilized in the complete DifFIQA approach.

\begin{figure}[t]
    \centering 
    \includegraphics[width=0.99\linewidth]{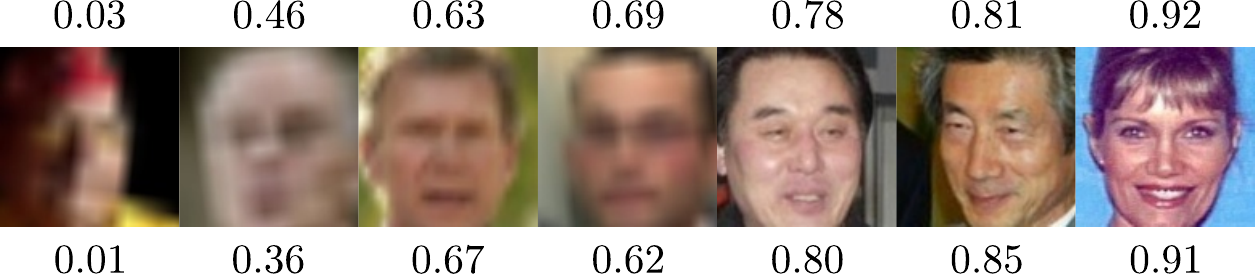}\vspace{1mm}
    \caption{\textbf{Illustration of the quality scores produced by the proposed FIQA techniques.} The scores on the top shows results for DifFIQA and the scores at the bottom for DifFIQA(R). While the concrete scores differ, both models generate similar rankings.\vspace{-3mm}}
    \label{fig:rankings}
\end{figure}

\subsection{Qualitative Evaluation}\label{sec:experiments_results:distributions}

While the proposed DifFIQA approach has a sound theoretical basis that links the forward and backward diffusion processes to face image quality, the distilled variant abstracts this relation away and approaches the FIQA task from a pure learning perspective. To get better insight into the characteristics of both models, we investigate in this section their behavior in a qualitative manner. \vspace{0.8mm} 

\noindent \textbf{Quality-score distributions.} In Figure~\ref{fig:distributions}, we compare the quality-score distributions, generated by DifFIQA and DifFIQA(R) on all seven test datasets. As can be seen, the two models produce very similar distributions, with a slight preference of DifFIQA(R) towards higher quality scores. 
\vspace{0.8mm} 

\noindent \textbf{Visual analysis.} In Figure \ref{fig:rankings}, we show example images from the XQLFW dataset and the corresponding quality scores, generated by the DifFIQA and DifFIQA(R) techniques. Note that both approaches produce a similar ranking but differ in the concrete quality score assigned to a given image. It is interesting to see that some blurry images with low (human-perceived) visual quality receive relatively high quality scores, as they feature frontal faces that may still be useful for recognition purposes. Additional qualitative results that illustrate the capabilties of the  DifFIQA model are also shown on the right part of Figure \ref{fig:teaser}. 
\vspace{-0.8mm}

\section{Conclusion}\label{sec:conclusion}

We have presented a novel approach to face image quality assessment (FIQA), called DifFIQA, that uses denoising diffusion probabilistic models as the basis for quality estimation. Through comprehensive experiments on multiple datasets we showed that the proposed model yields highly competitive results, when benchmarked against state-of-the-art techniques from the literature and that the runtime performance can be reduced significantly if the model is distilled into a quality predictor through a regression-based procedure. As part of our future work, we plan to investigate  extensions to our model, including transformer-based UNet alternatives and latent diffusion processes.

{\small
\bibliographystyle{ieee_fullname}
\bibliography{egbib}
}

\newpage

\section{Supplementary Material}

   In the main part of the paper, we evaluated the proposed DifFIQA technique in comprehensive experiments across $7$ diverse datasets, in comparison to $10$ state-of-the-art (SOTA) competitors, and with $4$ different face recognition models. In this supplementary material, we now show additional results using the same setup as in the main part of the paper that: (1) illustrate the performance of the model at another discard rate, (2) show the average performance of the proposed approach across all datasets and FR models and in comparison to  all considered SOTA techniques for two different discard rates, and (3) provide details on the runtime complexity of the DifFIQA model. Additionally, we also discuss the limitation of the proposed FIQA models and provide information on the reproducibility of the experiments described in the main part of the paper.

\begin{table}[!htb!]
    \centering
    \caption{\textbf{Comparison to the state-of-the-art.} The table reports pAUC scores at a discard rate of $0.2$ and a FMR of $10^{-3}$. Average results across all datasets are marked $\overline{\text{pAUC}}$. The best result for each dataset is shown in bold, the overall best result is colored green, the second-best blue and the third-best red. \vspace{-3mm}}
    \resizebox{\columnwidth}{!}{%
    \begin{tabular}{l | l l l l l l l | r}

\toprule
\multicolumn{9}{c}{\textbf{AdaFace - pAUC@FMR=}$10^{-3}$ ($\downarrow$)} \\
\midrule
\textbf{FIQA model} & \textbf{Adience} & \textbf{CALFW} & \textbf{CFP-FP} & \textbf{CPLFW} & \textbf{IJB-C} & \textbf{LFW} & \textbf{XQLFW} & $\overline{\text{pAUC}}$ \\ 
\cmidrule{1-9}
FaceQnet \cite{faceqnet}&  $0.969$ &  $0.960$ &  $0.772$ &  $0.935$ &  $1.133$ &  $0.934$ &  $0.969$ &  $0.953$ \\
SDD-FIQA \cite{sdd-fiqa}&  $0.884$ &  $0.911$ &  $0.632$ &  $0.789$ &  $0.854$ &  $0.857$ &  $0.907$ &  $0.833$ \\
PFE \cite{pfe}&  $0.873$ &  $0.917$ &  $0.659$ &  $0.772$ &  $0.918$ &  $0.854$ &  $0.885$ &  $0.840$ \\
PCNet \cite{pcnet} &  $1.003$ &  $0.985$ &  $0.893$ &  $0.926$ &  $0.843$ &  $0.730$ &  $0.999$ &  $0.911$ \\
MagFace \cite{magface}&  $0.890$ &  $0.900$ &  $0.632$ &  $0.747$ &  $0.915$ &  $0.735$ &  $0.958$ &  $0.825$ \\
LightQNet \cite{lightqnet}&  $0.890$ &  $0.925$ &  $0.711$ &  $0.784$ &  $0.846$ &  $0.837$ &  $0.836$ &  $0.833$ \\
SER-FIQ \cite{ser-fiq}&  $\mathbf{0.871}$ &  $0.930$ &  $0.563$ &  $0.715$ &  $0.812$ &  $0.982$ &  $n/a$ &  $0.812$ \\
FaceQAN \cite{faceqan}&  $0.905$ &  $0.942$ &  $\mathbf{0.474}$ &  $0.700$ &  $0.800$ &  $\mathbf{0.721}$ &  $0.764$ &  \cellcolor{blue!10}$\mathbf{0.758}$ \\
CR-FIQA \cite{cr-fiqa}&  $0.890$ &  $\mathbf{0.887}$ &  $0.504$ &  $\mathbf{0.684}$ &  $0.796$ &  $0.755$ &  $0.830$ &  \cellcolor{red!10}$\mathbf{0.764}$ \\
FaceQgen \cite{faceqgen}&  $0.889$ &  $0.967$ &  $0.774$ &  $0.778$ &  $0.877$ &  $0.887$ &  $0.814$ &  $0.855$ \\
\midrule
DifFIQA &  $0.897$ &  $0.932$ &  $0.500$ &  $0.698$ &  $0.813$ &  $0.770$ &  $0.769$ &  $0.768$ \\
DifFIQA(R) &  $0.893$ &  $0.913$ &  $0.505$ &  $0.696$ &  $\mathbf{0.796}$ &  $0.752$ &  $\mathbf{0.754}$ &  \cellcolor{green!10}$\mathbf{0.758}$ \\

\toprule
\multicolumn{9}{c}{\textbf{ArcFace - pAUC@FMR=}$10^{-3}$ ($\downarrow$)} \\
\midrule
\textbf{FIQA model} & \textbf{Adience} & \textbf{CALFW} & \textbf{CFP-FP} & \textbf{CPLFW} & \textbf{IJB-C} & \textbf{LFW} & \textbf{XQLFW} & $\overline{\text{pAUC}}$ \\ 
\cmidrule{1-9}
FaceQnet \cite{faceqnet} &  $0.957$ &  $0.970$ &  $0.761$ &  $0.918$ &  $1.123$ &  $0.934$ &  $0.933$ &  $0.942$ \\
SDD-FIQA \cite{sdd-fiqa}&  $0.841$ &  $0.931$ &  $0.637$ &  $0.829$ &  $0.806$ &  $0.857$ &  $0.874$ &  $0.825$ \\
PFE \cite{pfe}&  $\mathbf{0.823}$ &  $0.943$ &  $0.624$ &  $0.833$ &  $0.844$ &  $0.854$ &  $0.746$ &  $0.810$ \\
PCNet \cite{pcnet}&  $1.013$ &  $0.998$ &  $0.910$ &  $0.809$ &  $0.770$ &  $\mathbf{0.697}$ &  $1.003$ &  $0.886$ \\
MagFace \cite{magface}&  $0.852$ &  $0.925$ &  $0.683$ &  $0.809$ &  $0.867$ &  $0.712$ &  $0.961$ &  $0.830$ \\
LightQNet \cite{lightqnet}&  $0.840$ &  $0.930$ &  $0.706$ &  $0.857$ &  $0.788$ &  $0.814$ &  $0.772$ &  $0.816$ \\
SER-FIQ \cite{ser-fiq}&  $0.840$ &  $0.934$ &  $0.508$ &  $0.797$ &  $0.732$ &  $0.982$ &  $n/a$ &  $0.798$ \\
FaceQAN \cite{faceqan}&  $0.850$ &  $0.957$ &  $\mathbf{0.470}$ &  $0.771$ &  $0.731$ &  $0.699$ &  $0.710$ &  \cellcolor{blue!10}$\mathbf{0.741}$ \\
CR-FIQA \cite{cr-fiqa}&  $0.861$ &  $\mathbf{0.912}$ &  $0.475$ &  $0.791$ &  $\mathbf{0.724}$ &  $0.732$ &  $0.764$ &  $0.751$ \\
FaceQgen \cite{faceqgen}&  $0.857$ &  $0.980$ &  $0.823$ &  $0.834$ &  $0.823$ &  $0.865$ &  $0.786$ &  $0.853$ \\
\midrule
DifFIQA &  $0.848$ &  $0.931$ &  $0.493$ &  $\mathbf{0.771}$ &  $0.743$ &  $0.759$ &  $0.696$ &  \cellcolor{red!10}$\mathbf{0.749}$ \\
DifFIQA(R) &  $0.840$ &  $0.920$ &  $0.484$ &  $0.772$ &  $0.732$ &  $0.752$ &  $\mathbf{0.688}$ &  \cellcolor{green!10}$\mathbf{0.741}$ \\

\toprule
\multicolumn{9}{c}{\textbf{CosFace - pAUC@FMR=}$10^{-3}$ $(\downarrow)$} \\
\midrule
\textbf{FIQA model} & \textbf{Adience} & \textbf{CALFW} & \textbf{CFP-FP} & \textbf{CPLFW} & \textbf{IJB-C} & \textbf{LFW} & \textbf{XQLFW} & $\overline{\text{pAUC}}$ \\ 
\cmidrule{1-9}
FaceQnet \cite{faceqnet}&  $0.962$ &  $0.970$ &  $0.761$ &  $0.917$ &  $1.139$ &  $0.934$ &  $0.933$ &  $0.945$ \\
SDD-FIQA \cite{sdd-fiqa}&  $0.873$ &  $0.931$ &  $0.637$ &  $0.832$ &  $0.806$ &  $0.857$ &  $0.874$ &  $0.830$ \\
PFE \cite{pfe}&  $\mathbf{0.856}$ &  $0.943$ &  $0.624$ &  $0.837$ &  $0.848$ &  $0.854$ &  $0.746$ &  $0.816$ \\
PCNet \cite{pcnet} &  $1.005$ &  $0.998$ &  $0.910$ &  $0.861$ &  $0.776$ &  $\mathbf{0.697}$ &  $1.003$ &  $0.893$ \\
MagFace \cite{magface}&  $0.882$ &  $0.925$ &  $0.683$ &  $0.808$ &  $0.875$ &  $0.712$ &  $0.961$ &  $0.835$ \\
LightQNet \cite{lightqnet}&  $0.880$ &  $0.930$ &  $0.706$ &  $0.855$ &  $0.787$ &  $0.814$ &  $0.772$ &  $0.821$ \\
SER-FIQ \cite{ser-fiq}&  $0.863$ &  $0.934$ &  $0.508$ &  $0.790$ &  $0.725$ &  $0.982$ &  $n/a$ &  $0.800$ \\
FaceQAN \cite{faceqan}&  $0.890$ &  $0.957$ &  $\mathbf{0.470}$ &  $0.759$ &  $0.741$ &  $0.699$ &  $0.710$ &  \cellcolor{blue!10}$\mathbf{0.747}$ \\
CR-FIQA \cite{cr-fiqa}&  $0.884$ &  $\mathbf{0.912}$ &  $0.475$ &  $0.778$ &  $0.734$ &  $0.732$ &  $0.764$ &  $0.754$ \\
FaceQgen \cite{faceqgen}&  $0.880$ &  $0.980$ &  $0.823$ &  $0.821$ &  $0.824$ &  $0.865$ &  $0.786$ &  $0.854$ \\
\midrule
DifFIQA &  $0.881$ &  $0.931$ &  $0.493$ &  $\mathbf{0.758}$ &  $0.738$ &  $0.759$ &  $\mathbf{0.696}$ &  \cellcolor{red!10}$\mathbf{0.751}$ \\
DifFIQA(R) &  $0.870$ &  $0.931$ &  $0.484$ &  $0.758$ &  $\mathbf{0.723}$ &  $0.759$ &  $\mathbf{0.696}$ &  \cellcolor{green!10}$\mathbf{0.746}$ \\

\toprule
\multicolumn{9}{c}{\textbf{CurricularFace - pAUC@FMR}=$10^{-3}$ ($\downarrow$)} \\
\midrule
\textbf{FIQA model} & \textbf{Adience} & \textbf{CALFW} & \textbf{CFP-FP} & \textbf{CPLFW} & \textbf{IJB-C} & \textbf{LFW} & \textbf{XQLFW} & $\overline{\text{pAUC}}$ \\ 
\cmidrule{1-9}
FaceQnet \cite{faceqnet}&  $0.941$ &  $0.964$ &  $0.692$ &  $0.914$ &  $1.139$ &  $0.960$ &  $0.990$ &  $0.943$ \\
SDD-FIQA \cite{sdd-fiqa}&  $0.838$ &  $0.932$ &  $0.556$ &  $0.802$ &  $0.806$ &  $0.865$ &  $0.867$ &  $0.810$ \\
{PFE} \cite{pfe}&  $\mathbf{0.815}$ &  $0.937$ &  $0.539$ &  $0.793$ &  $0.848$ &  $0.863$ &  $0.900$ &  $0.814$ \\
{PCNet} \cite{pcnet}&  $1.000$ &  $0.993$ &  $0.931$ &  $0.938$ &  $0.776$ &  $0.732$ &  $0.971$ &  $0.906$ \\
{MagFace} \cite{magface}&  $0.841$ &  $0.921$ &  $0.624$ &  $0.779$ &  $0.875$ &  $0.736$ &  $0.901$ &  $0.811$ \\
{LightQNet} \cite{lightqnet}&  $0.827$ &  $0.938$ &  $0.574$ &  $0.815$ &  $0.787$ &  $0.834$ &  $0.857$ &  $0.805$ \\
{SER-FIQ} \cite{ser-fiq}&  $0.832$ &  $0.926$ &  $0.493$ &  $0.747$ &  $0.725$ &  $0.986$ &  $n/a$ &  $0.784$ \\
{FaceQAN} \cite{faceqan}&  $0.843$ &  $0.948$ &  $0.453$ &  $0.736$ &  $0.730$ &  $\mathbf{0.713}$ &  $0.908$ &  \cellcolor{red!10}$\mathbf{0.762}$ \\
{CR-FIQA} \cite{cr-fiqa}&  $0.859$ &  $\mathbf{0.908}$ &  $\mathbf{0.428}$ &  $\mathbf{0.729}$ &  $0.734$ &  $0.746$ &  $0.902$ &  \cellcolor{green!10}$\mathbf{0.758}$ \\
{FaceQgen} \cite{faceqgen}&  $0.858$ &  $0.972$ &  $0.754$ &  $0.806$ &  $0.824$ &  $0.894$ &  $0.836$ &  $0.849$ \\
\midrule
{DifFIQA} &  $0.851$ &  $0.919$ &  $0.499$ &  $0.738$ &  $0.738$ &  $0.771$ &  $0.863$ &  $0.768$ \\
{DifFIQA(R)} &  $0.832$ &  $0.922$ &  $0.467$ &  $0.740$ &  $\mathbf{0.723}$ &  $0.764$ &  $0.883$ &  \cellcolor{blue!10}$\mathbf{0.762}$ \\

\bottomrule
\multicolumn{9}{l}{$^\dagger$SER-FIQ was used to create XQLFW, so the results here are not reported for a fair comparison.}\vspace{-8mm}
    \end{tabular}
    }
    \label{tab:pauc_values}
\end{table}

\begin{table*}[!t!]
    \centering
    \caption{\textbf{Average performance over all seven test datasets and four FR models at a drop rate of $\mathbf{0.2}$.} The results are reported in terms of average pAUC score at the FMR of $10^{-3}$. The proposed DifFIQA(R) approach is overall the best performer. The best result is colored green, the second-best blue and the third-best red.\vspace{1mm}}
    \resizebox{0.9\textwidth}{!}{%
    \begin{tabular}{l l l l l l l l l l | l l}
    \toprule
    FaceQnet \cite{faceqnet} & SDD-FIQA \cite{sdd-fiqa} & PFE \cite{pfe} & PCNet \cite{pcnet} & MagFace \cite{magface} & LightQNet \cite{lightqnet} & SER-FIQ \cite{ser-fiq} & FaceQAN \cite{faceqan} & CR-FIQA \cite{cr-fiqa} & FaceQgen \cite{faceqgen} & DifFIQA & DifFIQA(R) \\
    \midrule
    $0.9458$ & $0.8244$ & $0.8197$ & $0.8989$ & $0.8253$ & $0.8183$ & $0.7985$ & \cellcolor{blue!10}$\mathbf{0.7519}$ & \cellcolor{red!10}$\mathbf{0.7567}$ & $0.8527$ & $0.7591$ & \cellcolor{green!10}$\mathbf{0.7518}$ \\ 
    \bottomrule
    \end{tabular}
    }
    \label{tab:avg_02}
\end{table*}

\begin{table*}[!t]
    \centering
    \caption{\textbf{Average performance over all seven test datasets and four FR models at a drop rate of $\mathbf{0.3}$.} The results are reported in terms of average pAUC score at the FMR of $10^{-3}$. The proposed DifFIQA(R) approach is overall the best performer. The best result is colored green, the second-best blue and the third-best red.\vspace{1mm}}
    \resizebox{0.9\textwidth}{!}{%
    \begin{tabular}{l l l l l l l l l l | l l}
    \toprule
    FaceQnet \cite{faceqnet} & SDD-FIQA \cite{sdd-fiqa} & PFE \cite{pfe} & PCNet \cite{pcnet} & MagFace \cite{magface} & LightQNet \cite{lightqnet} & SER-FIQ \cite{ser-fiq} & FaceQAN \cite{faceqan} & CR-FIQA \cite{cr-fiqa} & FaceQgen \cite{faceqgen} & DifFIQA & DifFIQA(R) \\
    \midrule
    $0.9315$ & $0.7483$ & $0.7497$ & $0.8691$ & $0.7635$ & $0.7412$ & $0.7292$ & $0.6847$ & \cellcolor{blue!10}$\mathbf{0.6800}$ & $0.7954$ & \cellcolor{red!10}$\mathbf{0.6822}$ & \cellcolor{green!10}$\mathbf{0.6768}$ \\ 
    \bottomrule
    \end{tabular}
    }
    \label{tab:avg_03}
\end{table*}

\begin{table*}[!htb]
    \centering
    \caption{\textbf{Detailed analysis of the runtime performance of DifFIQA in $\mathbf{ms}$.} The reported results were computed over the entire XQLFW dataset and for each component of the model separately. For DifFIQA the times are presented separately for the initialization $t_i$, the forward process $t_f$, the backward process $t_b$, embedding of the images $t_{fr}$, and the quality calculation $t_q$ steps. The symbol $\Sigma$ denotes the overall runtime. \vspace{1mm}}
    \resizebox{0.9\textwidth}{!}{%
    \begin{tabular}{ l | ccccc|c   }
        \toprule
        \textbf{Model component runtime} & $t_i$ & $t_f$ & $t_b$ & $t_{fr}$ & $t_{q}$ & $\Sigma$ \\
        \midrule
         Runtime in ms ($\mu \pm \sigma$) & $0.166\pm0.006$ & $0.192\pm0.010$ & $842.041\pm9.068$ &  $66.224\pm0.689$ & $166.335\pm1.750$ & $1074.627\pm11.458$ \\
        \bottomrule
    \end{tabular}
    }
    \label{tab:time_complexity_DifFIQA}
\end{table*}

\subsection{Additional Results}\label{sec:introduction}

\noindent \textbf{Comparison to SOTA techniques.} In Table~\ref{tab:pauc_values}, we present additional comparisons to the ten state-of-the-art techniques already considered in the main part of the paper. However, here the results are reported for a lower drop rate of $0.2$. We note again that the performance of FIQA techniques is most relevant at lower drop rates, since this facilitates real-world applications, as also emphasized in \cite{eval1}.  

From the presented results, we observe that the distilled model, DifFIQA(R) yields the lowest average pAUC scores (computed over the seven test datasets), when used with the AdaFace, ArcFace and CosFace models. With the CurricularFace model, DifFIQA(R) is the runner-up with performance close to the best performing CR-FIQA technique. It is worth noting that among the tested methods, four FIQA techniques performed significantly better than the rest across the four different FR models, i.e., CR-FIQA \cite{cr-fiqa}, FaceQAN \cite{faceqan} and the two diffusion-based models proposed in this paper, DifFIQA and DifFIQA(R). However, the distilled DifFIQA(R) technique is overall the top performer and fares particularly well on the most challenging datasets considered in the experiments, i.e., IJB-C and XQLFW. \vspace{0.8mm}     

\noindent \textbf{Overall performance.} To further illustrate the performance of the proposed DifFIQA and DifFIQA(R) techniques, we present in Tables \ref{tab:avg_02} and \ref{tab:avg_03} the average pAUC scores for two discard rates ($0.2$ and $0.3$), computed over the seven test datasets and all four considered FR models.  The reported results again support the findings already made above, i.e., FaceQAN, CR-FIQA, and our proposed techniques significantly outperform all other FIQA techniques, while DifFIQA(R) performs overall the best. \vspace{0.8mm}    

\noindent \textbf{Runtime complexity.} In the main part of the paper, we analyzed and tested all considered techniques from a runtime-performance perspective. Here, we explore the runtime complexity of DifFIQA in more detail to get better insight into the computationally most demanding steps of the approach. The whole method includes five steps: the  initialization step (i), which creates all the necessary image copies and converts them into tensors, the forward diffusion step (f), the backward diffusion step (b), the image embedding step (fr), and the quality score calculation step (q).  As can be seen from the reported results in Table \ref{tab:time_complexity_DifFIQA}, DifFIQA takes $1074$ms on average to estimate the quality of a single face image. Recall, that the distilled approach requires only around $1$ms for the same task. By far the most demanding part of the quality estimation procedure is the backward diffusion process, which iteratively denoises the given images, with an average time of a little more than $840$ms. Even though we use only $5$ iterations, we create for a single image $10$ noisy copies of the original and the flipped version. All of these images are then passed through the denoising network, which accounts for the high time complexity of the backward process. The generation of image embeddings also requires some time, i.e., $66$ms, as the step encapsulates the collection of all starting, noisy and reconstructed images into a single tensor as well as the forward pass through the FR model. In total, the image embedding steps need to produce embeddings for $60$ images, all constructed from the given input sample. 
The score computation also takes close to $170$ms, because it includes the calculation of five separate cosine similarities for all image copies, calculation of the average value over all copies and the data transfer from VRAM to RAM. 

\subsection{Limitations}\label{sec:experiments_results:limitations}

The proposed DDPM-based DifFIQA technique ensure highly competitive FIQA performance, but also has some \textbf{limitations}. One obvious limitation is the computational complexity that affects the model's runtime performance, as emphasized throughout the paper. While this can be addressed through a distillation procedure, the distillation process removes the relation between the (noising and denoising) tasks and image quality, and consequently impacts the interpretability of the results. From a conceptual point of view, the nosing and denoising steps probe the quality of the facial images by (in a sense) first obscuring important facial features and then measuring the ability to restore the obscured features through denoising. Such restoration-based solutions may depend, to a significant degree, on the restoration model utilized, which in our case is a CNN-based UNet that implements the denoising diffusion. While such models are known to be able to capture local image characteristics very well, they may be less capable in capturing key global image properties, and we plan to explore transformer-based models in our future work to further improve on this limitation.      


\subsection{Reproduciblity}\label{sec:conclusion}

We would like to note that all of our experiments are fully reproducible. Most of the models used for the implementation and testing of DifFIQA and DifFIQA(R) are publicly available from the official repositories, while all others can be obtained by request from the authors, i.e.: %
\begin{itemize}
    \item AdaFace: \\
    {\footnotesize \url{https://github.com/mk-minchul/AdaFace}}
    \item ArcFace: \\
    {\footnotesize \url{https://github.com/deepinsight/insightface}}
    \item CosFace: \\
    {\footnotesize \url{https://github.com/deepinsight/insightface}}
    \item CurricularFace: \\
    {\footnotesize \url{https://github.com/HuangYG123/CurricularFace}}
    \item FaceQnet: \\
    {\footnotesize \url{https://github.com/javier-hernandezo/FaceQnet}}
    \item SDD-FIQA: \\
    {\footnotesize \url{https://github.com/Tencent/TFace/tree/quality}}
    \item PFE: \\
    {\footnotesize \url{https://github.com/seasonSH/Probabilistic-Face-Embeddings}}
    \item PCNet: \\
    {\footnotesize Requested from authors}
    \item MagFace: \\
    {\footnotesize \url{https://github.com/IrvingMeng/MagFace}}
    \item LightQNet: \\
    {\footnotesize \url{https://github.com/KaenChan/lightqnet}}
    \item SER-FIQ: \\
    {\footnotesize \url{https://github.com/pterhoer/FaceImageQuality}}
    \item FaceQAN: \\
    {\footnotesize \url{https://github.com/LSIbabnikz/FaceQAN}}
    \item FaceQgen: \\
    {\footnotesize \url{https://github.com/javier-hernandezo/FaceQgen}}
    \item CR-FIQA: \\
    {\footnotesize \url{https://github.com/fdbtrs/CR-FIQA}}
    \item Diffusion models: \\
    {\footnotesize \url{https://github.com/lucidrains/denoising-diffusion-pytorch}}

\end{itemize}

Additionally, we also plan to publicly release the DifFIQA source code, including all training and testing scripts, model design and learned weights, once the review procedure is completed.

\end{document}